\documentclass[conference]{IEEEtran}
\IEEEoverridecommandlockouts
\usepackage{cite}
\usepackage{amsmath,amssymb,amsfonts}
\usepackage{algorithmic}
\usepackage{graphicx}
\usepackage{textcomp}
\usepackage{xcolor}
\usepackage{booktabs}
\usepackage{hyperref}
\hypersetup{
    colorlinks=true,
    allcolors=blue,
}

\def\BibTeX{{\rm B\kern-.05em{\sc i\kern-.025em b}\kern-.08em
    T\kern-.1667em\lower.7ex\hbox{E}\kern-.125emX}}

\DeclareMathOperator*{\argmax}{arg\,max}
\DeclareMathOperator*{\expect}{\mathbb{E}}
\newcommand{\customfootnotetext}[2]{{%
  \renewcommand{\thefootnote}{#1}%
  \footnotetext[0]{#2}}}%

\begin{document}

\title{Robust Fine-Tuning of Vision-Language Models for Domain Generalization\\
}

\author{
\IEEEauthorblockN{Kevin Vogt-Lowell$^*$}
\IEEEauthorblockA{Artificial Intelligence Technology \\
MIT Lincoln Laboratory \\
kevin.vogt-lowell@ll.mit.edu}
\and
\IEEEauthorblockN{Noah Lee$^*$}
\IEEEauthorblockA{Homeland Sensors and Analytics \\
MIT Lincoln Laboratory \\
noah.lee@ll.mit.edu}
\and
\IEEEauthorblockN{Theodoros Tsiligkaridis}
\IEEEauthorblockA{Artificial Intelligence Technology \\
MIT Lincoln Laboratory \\
ttsili@ll.mit.edu}
\and
\IEEEauthorblockN{Marc Vaillant}
\IEEEauthorblockA{Homeland Sensors and Analytics \\
MIT Lincoln Lab\\
marc.vaillant@ll.mit.edu}
}

\maketitle
\customfootnotetext{$^*$}{These authors contributed equally.}

\begin{abstract}

Transfer learning enables the sharing of common knowledge among models for a variety of downstream tasks, but traditional methods suffer in limited training data settings and produce narrow models incapable of effectively generalizing under distribution shifts. Foundation models have recently demonstrated impressive zero-shot inference capabilities and robustness under distribution shifts. However, zero-shot evaluation for these models has been predominantly confined to benchmarks with simple distribution shifts, limiting our understanding of their effectiveness under the more realistic shifts found in practice. Moreover, common fine-tuning methods for these models have yet to be evaluated against vision models in few-shot scenarios where training data is limited. To address these gaps, we present a new recipe for few-shot fine-tuning of the popular vision-language foundation model CLIP and evaluate its performance on challenging benchmark datasets with realistic distribution shifts from the WILDS collection. Our experimentation demonstrates that, while zero-shot CLIP fails to match performance of trained vision models on more complex benchmarks, few-shot CLIP fine-tuning outperforms its vision-only counterparts in terms of in-distribution and out-of-distribution accuracy at all levels of training data availability. This provides a strong incentive for adoption of foundation models within few-shot learning applications operating with real-world data. Code is available at \url{https://github.com/mit-ll/robust-vision-language-finetuning}.

\end{abstract}

\section{Introduction}

In transfer learning, the knowledge acquired by a model trained to solve one task is leveraged to solve other different yet related tasks. Traditionally, the approach to transfer learning has involved the fine-tuning of a uni-modal, shared feature extractor using a large, annotated, task-specific dataset \cite{survey_transfer_learning}. While this method is simple to implement and works well if the training dataset is similar to the target dataset, the resulting models have limited flexibility. Additionally, these models perform poorly without access to a large amount of quality target labels, so successfully training models via transfer learning proves difficult in disciplines that often operate within limited data environments. In these situations, successful deployment of new systems can involve considerable amounts of time and resources spent on data collection.

However, over the last several years, artificial intelligence research has increasingly shifted toward the creation of larger, more flexible models capable of reuse in a variety of applications.  In particular, a new class of models known as foundation models, a term first-popularized by the Stanford Institute for Human-Centered AI, has begun to transform the way AI systems are built and developed \cite{bert, dalle, gpt3}. Foundation models are defined as large neural networks that are trained on broad, unlabelled datasets through self-supervision at scale. Importantly, the unlabelled datasets used to train these models constitute samples of many different data types (i.e. RGB, thermal, text, audio, etc.) originating from a variety of perceptual data sources. Through self-supervised training over such diverse source data, foundation models become powerful, robust, general-purpose engines capable of multi-modal information processing and adaptation to a wide variety of downstream tasks with far less source data \cite{opps_and_benefits_fms}.

Domain generalization refers to learning models that are capable of maintaining high levels of performance on unseen target domains, or data distributions. In most applications involving real-world data, there exist real-world distribution shifts associated with the target environments in which the models are deployed. This shift can lead to a significant disparity between the distribution of data used to fine-tune a model and the distribution in which that model is actually tested \cite{DG_survey}. As a result, traditional transfer learning solutions also underperform when applied to domain generalization problems. Research has mostly attempted to address this issue through experimentation involving different pre-training and data augmentation techniques, producing some improvement with methods like empirical risk minimization \cite{wiles2022, dign:2023, fourier:2021, Gulrajani:2021, teterwak:2023}. Yet, vision-language foundation models have been shown to possess significant robustness to distribution shifts. Given this robustness and their reduced fine-tuning needs, can these models overcome the limitations of transfer learning faced by vision-only models in regard to limited data scenarios and generalization under challenging distribution shifts?

In \cite{radford2021}, Radford et al. explored generalization and zero-shot transfer using the vision-language foundation model, CLIP (Contrastive Language-Image Pre-training) \cite{radford2021}. By leveraging natural language supervision during visual representation learning, CLIP achieves impressive zero-shot performance across a variety of benchmarks, demonstrating significant generalization capabilities. Notably, the paper highlights that the accuracy of zero-shot CLIP matches that of a pre-trained ResNet50 on ImageNet. These models were also evaluated on additional benchmarks representing ImageNet distribution shifts, and CLIP produced consistent zero-shot accuracy across all distribution shifts while ResNet50 could not. However, the ImageNet distributions tested represent relatively simple examples when compared to distributions encountered in real-world applications, like those provided by the WILDS collection \cite{wilds}. When deploying zero-shot CLIP on more complex benchmarks with realistic distribution shifts, we found that the model significantly under-performed compared to trained state-of-the-art (SoTA) vision-only models.

Many studies have been conducted to explore and improve fine-tuning strategies for large, pre-trained models like CLIP. While existing fine-tuning strategies can substantially improve performance on target distributions, they also significantly reduce the model's robustness to distribution shifts \cite{kumar2022}. Wortsman et al. explored methods to maintain CLIP's robustness during fine-tuning \cite{wortsman2021}, finding that the ensembling of weights from zero-shot and fine-tuned CLIP models improved both accuracy under distribution shifts and accuracy on target distributions. Wortsman et al. further explored weight ensembling strategies in \cite{pmlr-v162-wortsman22a}, improving robustness and inference accuracy by averaging the weights of CLIP models fine-tuned using different hyper-parameters. Xin Zhang et al. tried an entirely different approach for adapting CLIP to unseen target distributions \cite{zhang2020}, proposing a novel approach called Domain Prompt Learning which automatically generated tailored text prompts for CLIP by estimating domain-specific features from the target distribution. Yet these studies, like \cite{radford2021}, only present results on distribution shifts derived from ImageNet, and they don't evaluate model performances in terms of gains over typical vision-only models. Yang Shu et al. \cite{shu2023} evaluate a fine-tuning strategy employing margin-based cross-entropy loss and beta moving average to improve generalization for CLIP, but add only DomainBed \cite{domainbed} as a benchmark alongside ImageNet. Furthermore, none of these studies evaluated their fine-tuning strategies in limited data environments.

In the field of remote sensing, deep learning models often operate within limited data environments. Smaller amounts of training data are available for these models due to a lack of raw data, increased difficulty of dataset annotation, and limitations of sensor characteristics. As a result, many studies on remote sensing image interpretation have explored few-shot learning methods in an effort to leverage the benefits of deep learning \cite{remote_sensing_survey}. While these studies have explored data augmentation and prior-knowledge based transfer learning approaches, none have evaluated foundation models as potential few-shot learning solutions. 

In this paper, we explore the use of vision-language foundation models as a modern solution to the limited training data and domain generalization issues inherent to previous transfer learning approaches. We provide the following contributions:

\begin{enumerate}
    \item We reveal that the zero-shot classification performance of CLIP does not match that of a fine-tuned SoTA vision-only model under challenging realistic distribution shifts.
    \item We demonstrate the superior performance of few-shot CLIP over a few-shot vision-only model in limited data environments containing realistic distribution shifts. We verify these results extend at various levels of training data availability by evaluating both in-distribution (ID) and out-of-distribution (OOD) robustness.
    \item We present a fine-tuning strategy for CLIP that combines cross-entropy training and stochastic weight averaging to further improve out-of-distribution robustness.
\end{enumerate}

\section{Methodology}

Our experiments compare results for both vision-only and vision-language models. Vision-language models are models where the visual output (i.e. image classification, object detection, segmentation) is conditioned on an additional text input. We adopt CLIP \cite{radford2021} as our model of choice to experiment with vision-language models.

\subsection{Vision-Language Models}
CLIP models are pre-trained with the task of matching text captions to images. The key advantages of CLIP as a research platform lie in its modularity, zero-shot capability, and simple formulation. Its encoders can be swapped out depending on desired accuracy versus model size trade-off, and the resulting performance scales predictably. Thus, CLIP is commonly used in systems as a foundation model: once pre-trained, CLIP can serve as the backbone of models fine-tuned for other common vision tasks. A key result of CLIP conditioning prediction on encoded text instead of predefined labels is the ability to transfer well to other tasks without fine-tuning (called zero-shot transfer). In \cite{radford2021}, the authors show that zero-shot CLIP outperforms a trained ResNet50-based \cite{he:2016} logistic regression classifier on 16 lower-complexity datasets. 

CLIP is implemented as two encoders, one for images and one for text. In \eqref{eq:1}, $I: Images \xrightarrow{} U$ is the image encoder, $T: Text \xrightarrow{} V$ is the text encoder. The encoder outputs are each projected to the embedding space $W$ by $\mathcal{P}_{I}: U\xrightarrow{}W$ and $\mathcal{P}_{T}: V\xrightarrow{}W$ respectively, and classification is determined by the cosine similarity ($S_{cos}(a, b) := a\cdot b / (\|a\|_2\|b\|_2)$) between each pair of text and image embeddings as shown in the below equation. 

\begin{equation}\label{eq:1}
\argmax_{y\in Text} S_{cos}(\mathcal{P}_I(I(x)),\mathcal{P}_T(T(y))    
\end{equation}

As a training objective, CLIP uses an InfoNCE loss function with temperature scaling ($\tau$), popularized by van den Oord et al. \cite{oord2018} and adapted for image-text learning by Zhang et al. \cite{zhang2020}. For a batch size $|B|$, and a given image sample and embedding $i_k$, we compute the cross-entropy of the sample with the aligned text $t_k$ and other unaligned text samples $t_{j\neq k}$ in the batch. The loss for a given text sample is computed in a similar yet symmetric manner. Losses of all image and text samples in the batch are averaged for back-propagation.

\begin{equation}\label{eq:2}
\mathcal{L}_{\text{image}} = - \expect_{k\in |B|}\begin{bmatrix}\log\dfrac{\exp(S_{cos}(i_k, t_k) / \tau)}{\sum_{t_j \in T} \exp(S_{cos}(i_k, t_j)/\tau)}\end{bmatrix}
\end{equation}
\begin{equation}\label{eq:3}
\mathcal{L}_{\text{text}} = - \expect_{k\in |B|}\begin{bmatrix}\log\dfrac{\exp(S_{cos}(i_k, t_k) / \tau)}{\sum_{i_j \in I} \exp(S_{cos}(i_j, t_k)/\tau)}\end{bmatrix}
\end{equation}

\subsection{Vision-Only Models}
For vision-only implementations of models, we extract the image encoder from CLIP. A randomly initialized trainable linear layer is added to the image encoder, with output dimension equal to the number of classes for the task we are fine-tuning for. Classification is performed using the softmax function, and training uses the standard cross-entropy objective. Following convention from \cite{radford2021}, we differentiate between vision-only models depending on whether the image encoder's weights are frozen or not. If the encoder's weights are frozen and only the classification layer is left trainable, we call those `linear probes'. Else, we label the models as `fine-tuned'.

\subsection{Weight Space Averaging}
Recent work has demonstrated that averaging the weights of multiple models can approximate the generalization capabilities of deep ensembles \cite{pmlr-v162-wortsman22a, ramé2023diverse, cha2021swad, Izmailov2018AveragingWL}, as well as provide additional robustness to distribution shift \cite{wortsman2021}. To investigate this idea, we implement SWA from \cite{Izmailov2018AveragingWL}. Our implementation runs for the final $n_\text{SWA} = 10$ epochs of training. At the end of each SWA epoch ($n_\text{cur} = 0, 1,\ldots, n_\text{SWA}$), the weights of the current model are aggregated with the moving average of model weights from past SWA epochs \eqref{eq:4}, which adds minimal overhead to the total training time.
\begin{equation}
\label{eq:4}
w_{\text{SWA}} \xleftarrow{} \dfrac{w_{\text{SWA}} \cdot n_\text{cur} + w_\text{current}}{n_\text{cur} + 1}
\end{equation}
During each epoch, the learning rate for each weight update per mini-batch follows a cosine annealing schedule \eqref{eq:5}, where $\eta_\text{min}$ and $\eta_\text{max}$ are the lower and upper bound learning rates and $T_i$ and $T_\text{max}$ are the current step and number of mini-batches in the epoch.
\begin{equation}
\label{eq:5}
\eta_{i+1}=\eta_\text{min}+\frac{1}{2}(\eta_\text{max}-\eta_\text{min})\begin{pmatrix}1+\cos\begin{pmatrix}\dfrac{T_i}{T_\text{max}}\cdot\pi\end{pmatrix}\end{pmatrix}
\end{equation}

\subsection{Distributed Training}
We also evaluate the effects of scaling the training of vision-language models. To successfully implement scaling for training CLIP, we make use of data parallelism \cite{li2020torchdistributed} to split the training dataset across nodes. Communication collectives via NCCL \cite{nccl} are used to synchronize gradients for models on different nodes. We also modify the training recipe of CLIP models by including the linear learning rate scaling rule and warmup periods suggested by Goyal et al. in \cite{goyal2017}. For our distributed training environment, we run on the TX-GAIA system at the MIT Lincoln Laboratory Supercomputing Center \cite{reuther2018interactive}. The system allows the use of 2 Intel Xeon Gold 6248 CPUs (40 total cores) and 2 NVIDIA Volta V100 GPUs (32 GB RAM each) per node. We launch jobs using a SLURM workload manager \cite{slurm}, starting from 1 V100 and scaling up to 32 V100 GPUs.

\section{Experimental Results}

This section presents the key findings from our experimental results. We begin by providing a brief overview of the datasets used, as well as relevant details regarding our experimental setup and hyper-parameter testing. Then, we show zero-shot CLIP's failure to match the in-distribution (ID) and out-of-distribution (OOD) performance of a pre-trained Vision Transformer (ViT) in domain generalization tasks with challenging distribution shifts. Next, we demonstrate that CLIP outperforms the ViT at all levels of training data availability above zero-shot in these same domain generalization tasks, but particularly within few-shot scenarios. We show these results extend across both ID and OOD testing, verifying CLIP's improved robustness to challenging distribution shifts over vision-only models in limited data environments. Finally, we conclude with an evaluation of the performance improvements provided by SWA and the training time speed-ups gained from scaling training to multiple GPUs.

\begin{figure*}[t]
    \centerline{\includegraphics[width=\textwidth]{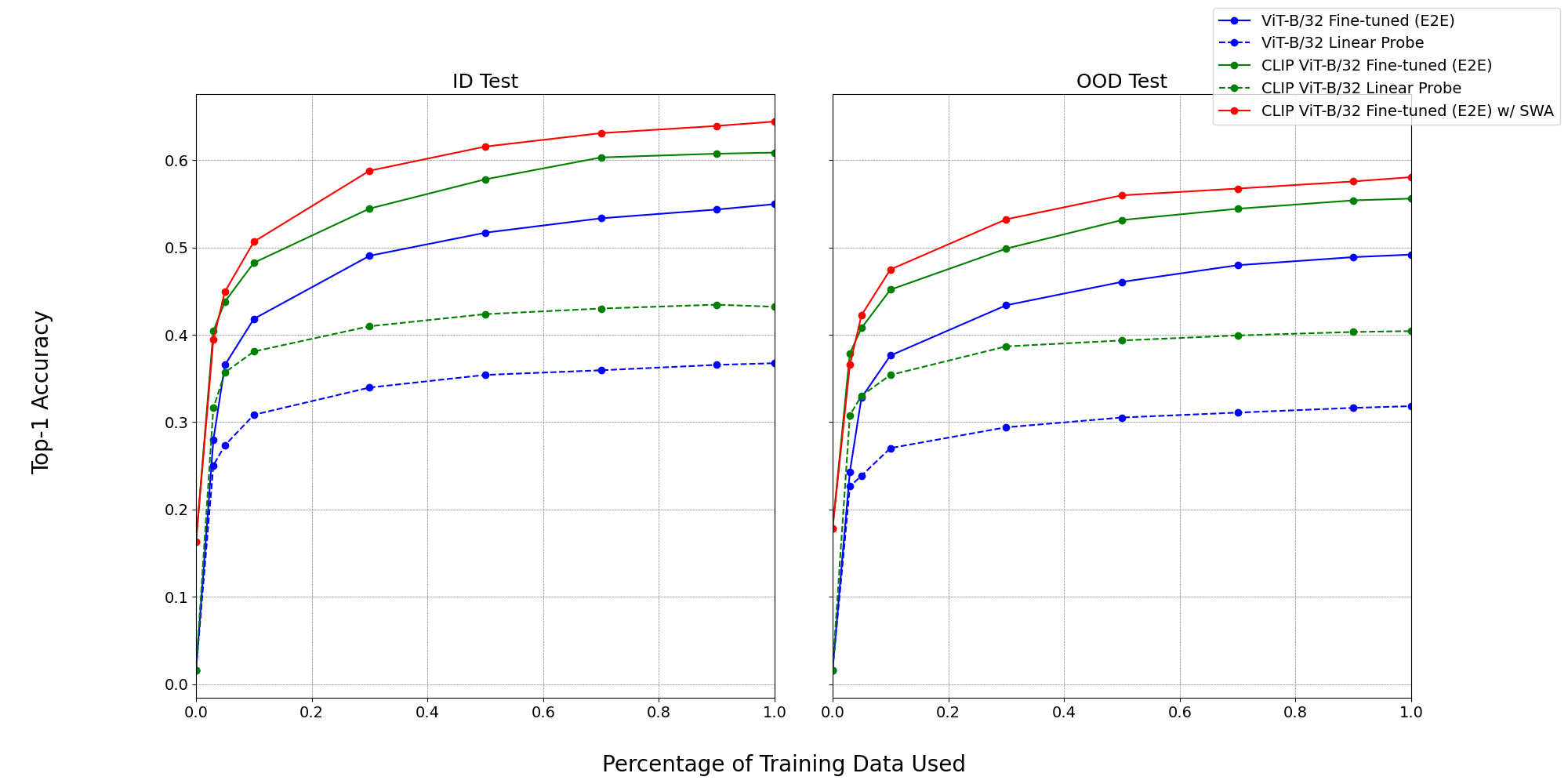}}
    \caption{Top-1 accuracies obtained on the FMoW in-distribution (ID Test) and out-of-distribution (OOD Test) test splits at different training data availabilities.
    }
    \label{fmow_vit_res_fig}
\end{figure*}

\begin{table*}
\setlength\tabcolsep{4.1pt}
\small
\caption{\label{tab:fmow_topk} FMoW Top-1 Accuracies}

\begin{center}
\begin{tabular}{l|ccccccccccc}
\toprule
{} &  &  \multicolumn{9}{c}{\textbf{Training Data Used}} \\
{} &  & 0\% & 3\%  &  5\%  &  10\%  &  30\%  &  50\%  &  70\%  &  90\% &  100\% \\
\midrule
\textbf{In-Distribution} &  & & & & & & & & & & \\
\quad ViT-B/32 Linear Probe & & 1.6 & 25.1{\scriptsize $\pm$0.3} & 27.4{\scriptsize $\pm$0.1} & 31.8{\scriptsize $\pm$0.1} & 34.0{\scriptsize $\pm$0.2} & 35.4{\scriptsize $\pm$0.4} & 35.9{\scriptsize $\pm$0.1} & 36.6{\scriptsize $\pm$0.1} & 36.7{\scriptsize $\pm$0.1} \\
\quad CLIP ViT-B/32 Linear Probe & & 1.6 & 31.6{\scriptsize $\pm$0.2} & 35.7{\scriptsize $\pm$0.2} & 38.1{\scriptsize $\pm$0.1} & 41.0{\scriptsize $\pm$0.1} & 42.4{\scriptsize $\pm$0.1} & 43.0{\scriptsize $\pm$0.2} & 43.4{\scriptsize $\pm$0.1} & 43.2{\scriptsize $\pm$0.1} \\
\quad ViT-B/32 Fine-tuned (E2E) & & 1.6 & 28.0{\scriptsize $\pm$2.5} & 36.5{\scriptsize $\pm$0.7} & 41.8{\scriptsize $\pm$0.3} & 49.0{\scriptsize $\pm$0.5} & 51.7{\scriptsize $\pm$0.2} & 53.3{\scriptsize $\pm$0.1} & 54.3{\scriptsize $\pm$0.1} & 55.0{\scriptsize $\pm$0.5} \\
\quad CLIP ViT-B/32 Fine-tuned (E2E) & & 16.3 & \textbf{40.4}{\scriptsize $\pm$0.5} & 43.8{\scriptsize $\pm$0.8} & 48.2{\scriptsize $\pm$0.6} & 54.4{\scriptsize $\pm$0.4} & 57.8{\scriptsize $\pm$0.3} & 60.3{\scriptsize $\pm$0.4} & 60.7{\scriptsize $\pm$0.4} & 60.9{\scriptsize $\pm$0.1}  \\
\quad CLIP ViT-B/32 Fine-tuned (E2E) w/ SWA & & 16.3 & 39.5{\scriptsize $\pm$1.2} & \textbf{44.9}{\scriptsize $\pm$0.5} & \textbf{50.7}{\scriptsize $\pm$0.4} & \textbf{58.8}{\scriptsize $\pm$0.5} & \textbf{61.5}{\scriptsize $\pm$0.1} & \textbf{63.1}{\scriptsize $\pm$0.3} & \textbf{63.9}{\scriptsize $\pm$0.4} & \textbf{64.4}{\scriptsize $\pm$0.1} \\
\\
\textbf{Out-of-Distribution} & & & & & & & & & \\
\quad ViT-B/32 Linear Probe & & 1.6 & 
22.7{\scriptsize $\pm$0.3} & 23.9{\scriptsize $\pm$0.1} & 27.0{\scriptsize$\pm$0.1} & 29.4{\scriptsize $\pm$0.1} & 30.5{\scriptsize$\pm$0.1} & 31.1{\scriptsize$\pm$0.1} & 31.6{\scriptsize$\pm$0.1} & 31.8{\scriptsize$\pm$0.1} \\
\quad CLIP ViT-B/32 Linear Probe & & 1.6 & 
30.8{\scriptsize $\pm$0.6} & 33.0{\scriptsize $\pm$0.5} & 35.4{\scriptsize $\pm$0.2} & 38.7{\scriptsize $\pm$0.0} & 39.3{\scriptsize$\pm$0.1} &39.9{\scriptsize $\pm$0.2} & 40.3{\scriptsize $\pm$0.1} & 40.4{\scriptsize $\pm$0.1} \\
\quad ViT-B/32 Fine-tuned (E2E) & & 1.6 & 
24.3{\scriptsize $\pm$2.2} & 32.8{\scriptsize $\pm$0.6} & 37.6{\scriptsize $\pm$0.2} & 43.4{\scriptsize $\pm$0.4} & 46.1{\scriptsize $\pm$0.1} & 48.0{\scriptsize $\pm$0.2} & 48.9{\scriptsize $\pm$0.5} & 49.2{\scriptsize $\pm$0.4} \\
\quad CLIP ViT-B/32 Fine-tuned (E2E) & & 17.8 & 
\textbf{37.9}{\scriptsize $\pm$0.6} & 40.8{\scriptsize $\pm$0.7} & 45.2{\scriptsize $\pm$0.6} & 49.9{\scriptsize $\pm$0.4} & 53.1{\scriptsize $\pm$0.2} & 54.4{\scriptsize $\pm$0.4} & 55.4{\scriptsize $\pm$0.3} & 55.6{\scriptsize $\pm$0.4}  \\
\quad CLIP ViT-B/32 Fine-tuned (E2E) w/ SWA & & 17.8 & 36.6{\scriptsize $\pm$0.8} & \textbf{42.2}{\scriptsize $\pm$0.9} & \textbf{47.5}{\scriptsize $\pm$0.1} & \textbf{53.2}{\scriptsize $\pm$0.4} & \textbf{56.0}{\scriptsize $\pm$0.2} & \textbf{56.7}{\scriptsize $\pm$0.1} & \textbf{57.6}{\scriptsize $\pm$0.5} & \textbf{58.0}{\scriptsize $\pm$0.1} \\

\bottomrule
\end{tabular}

\end{center}
\end{table*}

\subsection{Datasets}\label{datasets}

To evaluate model robustness and performance on realistic distribution shifts, our chosen datasets came from Stanford University's WILDS collection, a set of domain generalization benchmarks representing challenging distribution shifts faced in the wild.\cite{wilds} From WILDS, we selected the WILDS-FMoW (Functional Map of the World) \cite{fmow} and the WILDS-iWildCam datasets \cite{iwilds}, each containing geographic distribution shifts in satellite and wildlife imagery, respectively. Training and test distributions from the WILDS collection comprise disjoint sets of domains, allowing for adequate evaluation of model generalization to OOD test data.

\subsection{Experimental Setup and Hyper-Parameters}

For our experiments, we used a ViT-B/32 and a CLIP model using ViT-B/32 as its image backbone. To compare their performances at varying levels of data availability, the models were each trained on randomly sampled fractions of the training splits provided by FMoW and iWildCam. The exact fractions of available training data evaluated in these experiments can be seen in Table ~\ref{tab:fmow_topk}. Models were then evaluated on the entirety of the ID and OOD test splits from each of the respective datasets.

In each of these varied data experiments, ViT-B/32 and CLIP models were separately fine-tuned twice: once as linear probes and once end-to-end (E2E). By testing different fine-tuning strategies for each model, we evaluate the optimal strategy for each. This approach resulted in four model permutations for comparison: ViT-B/32 linear probe, ViT-B/32 fine-tuned, CLIP-ViT-B/32 linear probe, and CLIP-ViT-B/32 fine-tuned.

To account for the stochasticity inherent to the fine-tuning process, every model's top-1 accuracy (FMoW) or macro F1 score (iWildCam) was reported as the average of three separate, 20 epoch training runs. As part of model fine-tuning on each dataset, we included an exhaustive search over four learning rates of varying magnitudes: \{1e-2, 1e-3, 1e-4, 1e-5\}. All model permutations were fine-tuned using each of these learning rates in separate trials, and the best learning rate for each model selected based on the highest OOD performance for the given model configuration. The ID and OOD metrics reported in all figures and tables represent the averages yielded using each model permutation's best learning rate.

The magnitude of weight decay used during training on different datasets was chosen in a manner similar to the learning rate, examining \{1e-1, 1e-2, 1e-3, 1e-4\} as possible values. For both FMoW and iWildCam, a weight decay of 1e-4 was deemed appropriate. Batch size was held constant at a size of 128 across all experiments.

\begin{figure*}[ht!]
    \centerline{\includegraphics[width=\textwidth]{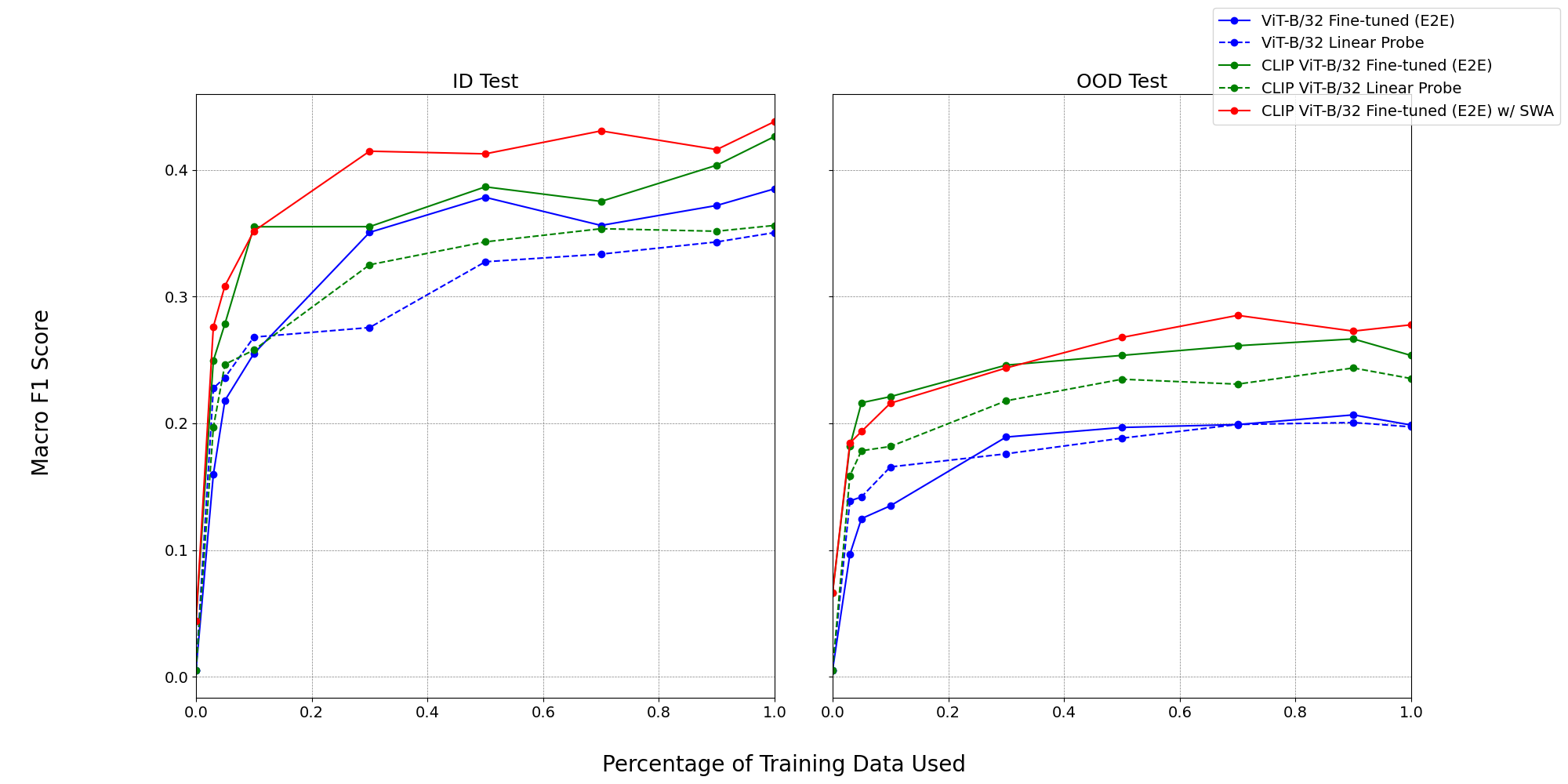}}
    \caption{Macro F1 scores obtained on the iWildCam in-distribution (ID Test) and out-of-distribution (OOD Test) test splits at different training data availabilities.}
    \label{iwildcam_vit_res_fig}
\end{figure*}

\begin{table*}[ht!]
\setlength\tabcolsep{4.1pt}
\small
\caption{\label{tab:iwildcam_f1} iWildCam Macro F1 Scores
}
\begin{center}
\begin{tabular}{l|ccccccccccc}
\toprule
{} &  &  \multicolumn{9}{c}{\textbf{Training Data Used}} \\
{} &  & 0\% & 3\%  &  5\%  &  10\%  &  30\%  &  50\%  &  70\%  &  90\% &  100\% \\
\midrule
\textbf{In-Distribution} &  & & & & & & & & & & \\
\quad ViT-B/32 Linear Probe & & 0.5 & 
22.8{\scriptsize $\pm$0.7} & 23.6{\scriptsize $\pm$0.6} & 26.8{\scriptsize $\pm$1.5} & 27.6{\scriptsize $\pm$0.2} & 32.8{\scriptsize $\pm$0.7} & 33.4{\scriptsize $\pm$0.8} & 34.3{\scriptsize $\pm$0.6} & 35.1{\scriptsize $\pm$0.6} \\
\quad CLIP ViT-B/32 Linear Probe & & 0.5 &
19.7{\scriptsize $\pm$0.1} & 24.7{\scriptsize $\pm$1.1} & 25.8{\scriptsize $\pm$1.0} & 32.5{\scriptsize $\pm$0.8} & 34.3{\scriptsize $\pm$1.4} & 35.4{\scriptsize $\pm$0.7} & 35.2{\scriptsize $\pm$0.2} & 35.6{\scriptsize $\pm$1.2} \\
\quad ViT-B/32 Fine-tuned (E2E) & & 0.5 & 
16.0{\scriptsize $\pm$0.6} & 21.8{\scriptsize $\pm$1.7} & 25.5{\scriptsize $\pm$0.1} & 35.1{\scriptsize $\pm$1.4} & 37.8{\scriptsize $\pm$1.7} & 35.6{\scriptsize $\pm$0.8} & 37.2{\scriptsize $\pm$0.4} & 38.5{\scriptsize $\pm$0.6} \\
\quad CLIP ViT-B/32 Fine-tuned (E2E) & & 4.1 &
25.0{\scriptsize $\pm$2.0} & 27.9{\scriptsize $\pm$1.6} & \textbf{35.5}{\scriptsize $\pm$1.1} & 35.5{\scriptsize $\pm$3.6} & 38.7{\scriptsize $\pm$3.7} & 37.5{\scriptsize $\pm$1.2} & 40.4{\scriptsize $\pm$0.5} & 42.6{\scriptsize $\pm$2.1}  \\
\quad CLIP ViT-B/32 Fine-tuned (E2E) w/ SWA & & 4.1 & 
\textbf{27.6}{\scriptsize $\pm$0.9} & \textbf{30.9}{\scriptsize $\pm$1.1} & 35.2{\scriptsize $\pm$0.9} & \textbf{41.5}{\scriptsize $\pm$0.9} & \textbf{41.3}{\scriptsize$\pm$1.4} & \textbf{43.1}{\scriptsize $\pm$0.9} & \textbf{41.6}{\scriptsize $\pm$0.2} & \textbf{43.8}{\scriptsize $\pm$0.5} \\
\\
\textbf{Out-of-Distribution} & & & & & & & & & \\
\quad ViT-B/32 Linear Probe & & 0.5 & 
13.9{\scriptsize $\pm$0.2} & 14.2{\scriptsize $\pm$0.3} & 16.6{\scriptsize$\pm$0.6} & 17.6{\scriptsize $\pm$0.3} & 18.8{\scriptsize$\pm$0.3} & 19.9{\scriptsize$\pm$0.3} & 20.1{\scriptsize$\pm$0.7} & 19.7{\scriptsize$\pm$0.2} \\
\quad CLIP ViT-B/32 Linear Probe & & 0.5 & 
15.9{\scriptsize $\pm$0.4} & 17.8{\scriptsize $\pm$0.6} & 18.2{\scriptsize $\pm$0.4} & 21.8{\scriptsize $\pm$0.7} & 23.5{\scriptsize$\pm$0.7} & 23.1{\scriptsize $\pm$0.4} & 24.4{\scriptsize $\pm$0.3} & 23.5{\scriptsize $\pm$0.9} \\
\quad ViT-B/32 Fine-tuned (E2E) & & 0.5 & 
9.7{\scriptsize $\pm$0.3} & 12.5{\scriptsize $\pm$1.6} & 13.5{\scriptsize $\pm$1.0} & 18.9{\scriptsize $\pm$0.6} & 19.7{\scriptsize $\pm$0.7} & 19.9{\scriptsize $\pm$0.3} & 20.7{\scriptsize $\pm$0.6} & 19.9{\scriptsize $\pm$0.7} \\
\quad CLIP ViT-B/32 Fine-tuned (E2E) & & 6.6 & 
18.2{\scriptsize $\pm$0.3} & \textbf{21.6}{\scriptsize $\pm$1.1} & \textbf{22.1}{\scriptsize $\pm$1.0} & \textbf{24.6}{\scriptsize $\pm$0.5} & 25.4{\scriptsize $\pm$1.3} & 26.1{\scriptsize $\pm$0.9} & 26.7{\scriptsize $\pm$1.0} & 25.4{\scriptsize $\pm$1.4}  \\
\quad CLIP ViT-B/32 Fine-tuned (E2E) w/ SWA & & 6.6 & 
\textbf{18.4}{\scriptsize $\pm$0.8} & 19.4{\scriptsize $\pm$0.7} & 21.6{\scriptsize $\pm$0.6} & 24.4{\scriptsize $\pm$0.4} & \textbf{26.8}{\scriptsize$\pm$0.7} & \textbf{28.5}{\scriptsize $\pm$0.7} & \textbf{27.3}{\scriptsize $\pm$0.8} & \textbf{27.8}{\scriptsize $\pm$0.8} \\

\bottomrule
\end{tabular}

\end{center}
\end{table*}

\subsection{CLIP in Limited Data Scenarios}

While zero-shot CLIP is capable of achieving comparable performance to vision-only linear probes on simpler distributions \cite{radford2021}, our results highlight that the zero-shot performance of CLIP does not match that of a fine-tuned vision-only linear probe when applied to benchmarks from WILDS with challenging and realistic distribution shifts. As seen in Table ~\ref{tab:fmow_topk}, zero-shot inference from CLIP scores a top-1 ID accuracy of 16.3\% and OOD accuracy of 17.8\% on FMoW, whereas the ViT-B/32 obtains an ID accuracy of 36.7\% and OOD accuracy of 31.8\% when trained as a linear probe using the full dataset. This trend holds for iWildCam as well, and can be seen in Table ~\ref{tab:iwildcam_f1}. However, the zero-shot accuracies from CLIP still lie far above the accuracies expected from random guessing (1.6\% and 0.5\%, respectively), indicating that pre-trained vision-language models have some semantic understanding that can be leveraged.

The top-1 accuracies on FMoW for the linear probe and E2E fine-tuned ViT-B/32 and CLIP model permutations across multiple levels of training data availability above zero-shot can be seen in Fig. ~\ref{fmow_vit_res_fig} and Table ~\ref{tab:fmow_topk}.  We can see that, at all levels of data availability, each of the CLIP model variations significantly outperforms its vision-only ViT-B/32 counterpart in terms of top-1 accuracy. This trend holds across testing on both ID and OOD samples. In particular, some of the biggest differences in performance between CLIP and ViT-B/32 can be seen in the few-shot scenarios, where data was limited. For example, the average difference between the top-1 accuracies for the E2E fine-tuned models was 6.1\% ID and 6.7\% OOD at training data availabilities above 5\%, but was 9.9\% ID and 10.8\% OOD at training data availabilities at and below 5\%. For all models, accuracy expectedly improves given increased amounts of available training data. The results for the same model permutations evaluated on iWildCam can be found in Figure ~\ref{iwildcam_vit_res_fig} and Table ~\ref{tab:iwildcam_f1}. Similar to the findings for FMoW, the CLIP model variants outperform the the ViT-B/32 at nearly all levels of training data availability when evaluated on both ID and OOD test samples, with the largest differences occurring when training data is most limited. Macro F1 scores tend to improve as models are given increased access to training data but some fluctuation occurs, likely related to the higher amount of rare class labels found in iWildCam compared to FMoW.  In addition, the results from both datasets demonstrate that E2E fine-tuning of either model architecture yields higher accuracies for both ID and OOD testing than those obtained through linear probing alone. By incorporating SWA into the fine-tuning process, we further improved outcomes on both datasets at nearly all levels of data availability.

In terms of model robustness, the CLIP model variants were more robust than their ViT-B/32 counterparts. For FMoW, the average drop in top-1 accuracy between the ID and OOD test results was 4.4\% for ViT-B/32 and 2.6\% for CLIP when linearly probed, and 4.9\% for ViT-B/32 and 4.3\% for CLIP when fine-tuned E2E. For iWildCam, we observe the same trends regarding robustness under distribution shifts: F1 scores drop by 11.6\% for ViT-B/32 and 9.4\% for CLIP when linearly probed, and 14.0\% for ViT-B/32 and 11.6\% for CLIP when fine-tuned E2E. As expected from \cite{kumar2022}, the differences in robustness between the ViT and CLIP were most pronounced in few-shot environments, yet robustness for CLIP diminished when fine-tuned with larger amounts of training data. While SWA improved performance metrics for our best CLIP models, robustness across distribution shifts remained approximately the same.

\subsection{Distributed Training}
We investigate the effects of scaling training with SWA to multiple GPUs for Vision-Language models. Our experiments run for 30 epochs on the FMoW dataset. Training includes a 5 epoch linear warmup period, and the learning rate is scaled $\eta_1 \xleftarrow{} k\cdot\eta_0$ by the number of GPUs used for training $k$. Note that scaling the number of GPUs is equivalent to scaling the effective batch size of training, since each GPU receives 128 samples per batch. 
\begin{table}[ht]
\label{fig:distributed_training}
\caption{Distributed training scalability on FMoW}
\centering
\begin{tabular}{r|ccl}
    \toprule
    {}& \multicolumn{2}{c}{\textbf{Top-1 Accuracy}} & {} \\
    \textbf{GPUs} & \textbf{ID} & \textbf{OOD} & \textbf{\% Scale Efficiency}\\
    \midrule
    1 & \textbf{61.0} $\pm 0.5$ & 55.1 $\pm0.9$ & 100.0 (367.21s)\\
    2 & 60.4 $\pm0.3$ & \textbf{55.6} $\pm 0.4$ & 85.97 (213.53s)\\
    4 & 58.9 $\pm0.4$ & 54.2 $\pm0.6$ & 81.75 (112.30s)\\
    8 & 58.8 $\pm0.6$ & 53.5 $\pm0.1$ & 76.53 (59.98s)\\
    16 & 59.6 $\pm0.7$ & 54.2 $\pm0.5$ & 78.98 (29.06s)\\
    32 & 55.7 $\pm1.6$ & 50.7 $\pm1.2$ & 61.46 (18.67s)\\
    \bottomrule
\end{tabular}
\end{table}
We report the Top-1 ID and OOD accuracy, as well as the scale efficiency, which we define as the ratio between the time per epoch on one GPU $T_1$ scaled to $k$ GPUs $T_1/k$ and the actual time per epoch training on $k$ GPUs $T_k$. As expected in \ref{fig:distributed_training}, the scale efficiency decreases as $k$ increases due to inter-process communication overheads, necessary synchronization during training, and serial execution during evaluation. 
\begin{figure}[ht]
    \centering
    \includegraphics[scale=0.5]{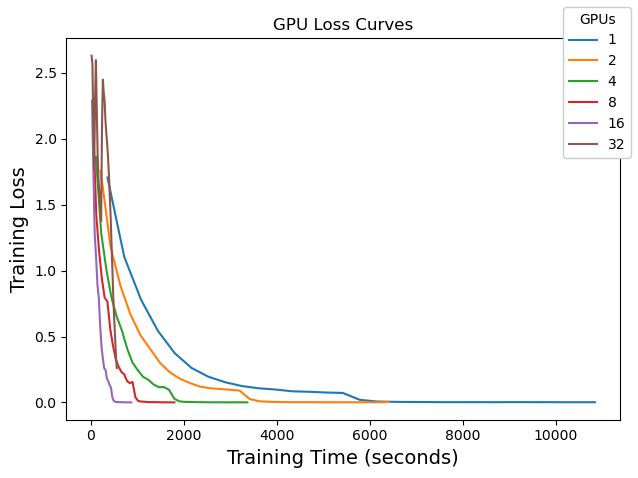}
    \caption{Training loss curves for models trained on varying numbers of GPUs.}
    \label{fig:dist_losses}
\end{figure}
We also see in \ref{fig:distributed_training} that OOD accuracy does not begin to significantly degrade until 32 GPUs are used. This matches with the results of \cite{goyal2017}, where using a linear warmup, they observe no degradation in performance training on ImageNet until reaching batch sizes greater than 8192. Our results are further validated by \ref{fig:dist_losses}, which shows that all training runs except for 32 GPUs converge to the same loss after 30 epochs, even though there is variation in the loss at the beginning of training. Overall, including the linear warmup helps make the training procedure robust to scaling.
\section{Conclusion}
This paper addresses the limitations of traditional transfer learning methods and the need for robust models capable of generalizing under distribution shifts. A new recipe for few-shot fine-tuning of the vision-language foundation model CLIP is proposed and its performance is evaluated on challenging benchmark datasets with realistic distribution shifts. The experimentation demonstrates that while CLIP's zero-shot inference capabilities do not match the performance of trained vision models on complex benchmarks, few-shot CLIP fine-tuning surpasses its vision-only counterparts in terms of in-distribution and out-of-distribution accuracy across different levels of training data availability. Adapting to limited training data, foundation models like CLIP offer enhanced performance and robustness in the face of distribution shifts, paving the way for more effective and practical machine learning applications. In future work, we will focus on improving domain generalization and novel class generalization performance of vision-language models in few-shot regimes by leveraging robust optimization and parameter-efficient methods.

\section*{Acknowledgment}

The authors acknowledge the MIT Lincoln Laboratory Supercomputing Center for providing the high performance computing resources that have contributed to the research results reported within this paper.

DISTRIBUTION STATEMENT A. Approved for public release. Distribution is unlimited. This material is based upon work supported by the Under Secretary of Defense for Research and Engineering under Air Force Contract No. FA8702-15-D-0001. Any opinions, findings, conclusions or recommendations expressed in this material are those of the author(s) and do not necessarily reflect the views of the Under Secretary of Defense for Research and Engineering.


\bibliographystyle{ieeetr}
\bibliography{bib/references}

\begin{thebibliography}{10}

\bibitem{survey_transfer_learning}
K.~Weiss, T.~Khoshgoftaar, and D.~Wang, ``A survey of transfer learning,'' {\em
  Journal of Big Data}, vol.~3, 05 2016.

\bibitem{bert}
J.~Devlin, M.~Chang, K.~Lee, and K.~Toutanova, ``{BERT:} pre-training of deep
  bidirectional transformers for language understanding,'' {\em CoRR},
  vol.~abs/1810.04805, 2018.

\bibitem{dalle}
A.~Ramesh, M.~Pavlov, G.~Goh, S.~Gray, C.~Voss, A.~Radford, M.~Chen, and
  I.~Sutskever, ``Zero-shot text-to-image generation,'' {\em CoRR},
  vol.~abs/2102.12092, 2021.

\bibitem{gpt3}
T.~B. Brown, B.~Mann, N.~Ryder, M.~Subbiah, J.~Kaplan, P.~Dhariwal,
  A.~Neelakantan, P.~Shyam, G.~Sastry, A.~Askell, S.~Agarwal,
  A.~Herbert{-}Voss, G.~Krueger, T.~Henighan, R.~Child, A.~Ramesh, D.~M.
  Ziegler, J.~Wu, C.~Winter, C.~Hesse, M.~Chen, E.~Sigler, M.~Litwin, S.~Gray,
  B.~Chess, J.~Clark, C.~Berner, S.~McCandlish, A.~Radford, I.~Sutskever, and
  D.~Amodei, ``Language models are few-shot learners,'' {\em CoRR},
  vol.~abs/2005.14165, 2020.

\bibitem{opps_and_benefits_fms}
R.~Bommasani, D.~A. Hudson, E.~Adeli, R.~B. Altman, S.~Arora, S.~von Arx, M.~S.
  Bernstein, J.~Bohg, A.~Bosselut, E.~Brunskill, E.~Brynjolfsson, S.~Buch,
  D.~Card, R.~Castellon, N.~S. Chatterji, A.~S. Chen, K.~Creel, J.~Q. Davis,
  D.~Demszky, C.~Donahue, M.~Doumbouya, E.~Durmus, S.~Ermon, J.~Etchemendy,
  K.~Ethayarajh, L.~Fei{-}Fei, C.~Finn, T.~Gale, L.~Gillespie, K.~Goel, N.~D.
  Goodman, S.~Grossman, N.~Guha, T.~Hashimoto, P.~Henderson, J.~Hewitt, D.~E.
  Ho, J.~Hong, K.~Hsu, J.~Huang, T.~Icard, S.~Jain, D.~Jurafsky, P.~Kalluri,
  S.~Karamcheti, G.~Keeling, F.~Khani, O.~Khattab, P.~W. Koh, M.~S. Krass,
  R.~Krishna, R.~Kuditipudi, and et~al., ``On the opportunities and risks of
  foundation models,'' {\em CoRR}, vol.~abs/2108.07258, 2021.

\bibitem{DG_survey}
K.~Zhou, Z.~Liu, Y.~Qiao, T.~Xiang, and C.~C. Loy, ``Domain generalization: {A}
  survey,'' {\em CoRR}, vol.~abs/2103.02503, 2021.

\bibitem{wiles2022}
O.~Wiles, S.~Gowal, F.~Stimberg, S.-A. Rebuffi, I.~Ktena, K.~D. Dvijotham, and
  A.~T. Cemgil, ``A fine-grained analysis on distribution shift,'' in {\em
  International Conference on Learning Representations}, 2022.

\bibitem{dign:2023}
T.~Tsiligkaridis and A.~Tsiligkaridis, ``Diverse gaussian noise consistency
  regularization for robustness and uncertainty calibration,'' in {\em IJCNN},
  2023.

\bibitem{fourier:2021}
R.~Soklaski, M.~Yee, and T.~Tsiligkaridis, ``Fourier-based augmentations for
  improved robustness and uncertainty calibration,'' in {\em NeurIPS 2021
  Workshop on Distribution Shifts}, 2021.

\bibitem{Gulrajani:2021}
I.~Gulrajani and D.~Lopez-Paz, ``In search of lost domain generalization,'' in
  {\em ICLR}, 2021.

\bibitem{teterwak:2023}
P.~Teterwak, K.~Saito, T.~Tsiligkaridis, K.~Saenko, and B.~Plummer, ``Erm++: An
  improved baseline for domain generalization,'' in {\em ICML 2023 Workshop on
  Spurious Correlations, Invariance and Stability}, 2023.

\bibitem{radford2021}
A.~Radford, J.~W. Kim, C.~Hallacy, A.~Ramesh, G.~Goh, S.~Agarwal, G.~Sastry,
  A.~Askell, P.~Mishkin, J.~Clark, {\em et~al.}, ``Learning transferable visual
  models from natural language supervision,'' in {\em International conference
  on machine learning}, pp.~8748--8763, PMLR, 2021.

\bibitem{wilds}
P.~W. Koh, S.~Sagawa, H.~Marklund, S.~M. Xie, M.~Zhang, A.~Balsubramani, W.~Hu,
  M.~Yasunaga, R.~L. Phillips, S.~Beery, J.~Leskovec, A.~Kundaje, E.~Pierson,
  S.~Levine, C.~Finn, and P.~Liang, ``{WILDS:} {A} benchmark of in-the-wild
  distribution shifts,'' {\em CoRR}, vol.~abs/2012.07421, 2020.

\bibitem{kumar2022}
A.~Kumar, A.~Raghunathan, R.~Jones, T.~Ma, and P.~Liang, ``Fine-tuning can
  distort pretrained features and underperform out-of-distribution,'' 2022.

\bibitem{wortsman2021}
M.~Wortsman, G.~Ilharco, M.~Li, J.~W. Kim, H.~Hajishirzi, A.~Farhadi,
  H.~Namkoong, and L.~Schmidt, ``Robust fine-tuning of zero-shot models,'' {\em
  CoRR}, vol.~abs/2109.01903, 2021.

\bibitem{pmlr-v162-wortsman22a}
M.~Wortsman, G.~Ilharco, S.~Y. Gadre, R.~Roelofs, R.~Gontijo-Lopes, A.~S.
  Morcos, H.~Namkoong, A.~Farhadi, Y.~Carmon, S.~Kornblith, and L.~Schmidt,
  ``Model soups: averaging weights of multiple fine-tuned models improves
  accuracy without increasing inference time,'' in {\em Proceedings of the 39th
  International Conference on Machine Learning} (K.~Chaudhuri, S.~Jegelka,
  L.~Song, C.~Szepesvari, G.~Niu, and S.~Sabato, eds.), vol.~162 of {\em
  Proceedings of Machine Learning Research}, pp.~23965--23998, PMLR, 17--23 Jul
  2022.

\bibitem{zhang2020}
Y.~Zhang, H.~Jiang, Y.~Miura, C.~D. Manning, and C.~P. Langlotz, ``Contrastive
  learning of medical visual representations from paired images and text,''
  {\em CoRR}, vol.~abs/2010.00747, 2020.

\bibitem{shu2023}
Y.~Shu, X.~Guo, J.~Wu, X.~Wang, J.~Wang, and M.~Long, ``Clipood: Generalizing
  clip to out-of-distributions,'' 2023.

\bibitem{domainbed}
I.~Gulrajani and D.~Lopez{-}Paz, ``In search of lost domain generalization,''
  {\em CoRR}, vol.~abs/2007.01434, 2020.

\bibitem{remote_sensing_survey}
X.~Sun, B.~Wang, Z.~Wang, H.~Li, H.~Li, and K.~Fu, ``Research progress on
  few-shot learning for remote sensing image interpretation,'' {\em IEEE
  Journal of Selected Topics in Applied Earth Observations and Remote Sensing},
  vol.~14, pp.~2387--2402, 2021.

\bibitem{he:2016}
K.~He, X.~Zhang, S.~Ren, and J.~Sun, ``Deep residual learning for image
  recognition,'' in {\em 2016 IEEE Conference on Computer Vision and Pattern
  Recognition (CVPR)}, pp.~770--778, 2016.

\bibitem{oord2018}
A.~van~den Oord, Y.~Li, and O.~Vinyals, ``Representation learning with
  contrastive predictive coding,'' {\em CoRR}, vol.~abs/1807.03748, 2018.

\bibitem{ramé2023diverse}
A.~Ramé, M.~Kirchmeyer, T.~Rahier, A.~Rakotomamonjy, P.~Gallinari, and
  M.~Cord, ``Diverse weight averaging for out-of-distribution generalization,''
  2023.

\bibitem{cha2021swad}
J.~Cha, S.~Chun, K.~Lee, H.-C. Cho, S.~Park, Y.~Lee, and S.~Park, ``Swad:
  Domain generalization by seeking flat minima,'' in {\em Advances in Neural
  Information Processing Systems (NeurIPS)}, 2021.

\bibitem{Izmailov2018AveragingWL}
P.~Izmailov, D.~Podoprikhin, T.~Garipov, D.~P. Vetrov, and A.~G. Wilson,
  ``Averaging weights leads to wider optima and better generalization,'' in
  {\em Conference on Uncertainty in Artificial Intelligence}, 2018.

\bibitem{li2020torchdistributed}
S.~Li, Y.~Zhao, R.~Varma, O.~Salpekar, P.~Noordhuis, T.~Li, A.~Paszke,
  J.~Smith, B.~Vaughan, P.~Damania, and S.~Chintala, ``Pytorch distributed:
  Experiences on accelerating data parallel training,'' 2020.

\bibitem{nccl}
{Nvidia}, ``Nvidia collective communication library.''

\bibitem{goyal2017}
P.~Goyal, P.~Doll{\'{a}}r, R.~B. Girshick, P.~Noordhuis, L.~Wesolowski,
  A.~Kyrola, A.~Tulloch, Y.~Jia, and K.~He, ``Accurate, large minibatch {SGD:}
  training imagenet in 1 hour,'' {\em CoRR}, vol.~abs/1706.02677, 2017.

\bibitem{reuther2018interactive}
A.~Reuther, J.~Kepner, C.~Byun, S.~Samsi, W.~Arcand, D.~Bestor, B.~Bergeron,
  V.~Gadepally, M.~Houle, M.~Hubbell, M.~Jones, A.~Klein, L.~Milechin,
  J.~Mullen, A.~Prout, A.~Rosa, C.~Yee, and P.~Michaleas, ``Interactive
  supercomputing on 40,000 cores for machine learning and data analysis,'' in
  {\em 2018 IEEE High Performance extreme Computing Conference (HPEC)},
  pp.~1--6, IEEE, 2018.

\bibitem{slurm}
A.~B. Yoo, M.~A. Jette, and M.~Grondona, ``Slurm: Simple linux utility for
  resource management,'' in {\em Job Scheduling Strategies for Parallel
  Processing} (D.~Feitelson, L.~Rudolph, and U.~Schwiegelshohn, eds.), (Berlin,
  Heidelberg), pp.~44--60, Springer Berlin Heidelberg, 2003.

\bibitem{fmow}
G.~A. Christie, N.~Fendley, J.~Wilson, and R.~Mukherjee, ``Functional map of
  the world,'' {\em CoRR}, vol.~abs/1711.07846, 2017.

\bibitem{iwilds}
S.~Beery, E.~Cole, and A.~Gjoka, ``The iwildcam 2020 competition dataset,''
  {\em CoRR}, vol.~abs/2004.10340, 2020.

\end{thebibliography}

\end{document}